\documentclass[11pt]{article}

\usepackage[preprint]{acl}

\usepackage{times}
\usepackage{latexsym}
\usepackage[T1]{fontenc}
\usepackage[utf8]{inputenc}
\usepackage{microtype}
\usepackage{graphicx}
\usepackage{booktabs}
\usepackage{multirow}
\usepackage{xspace}
\usepackage{xcolor}
\usepackage{amsmath}
\usepackage{amssymb}
\usepackage{placeins}
\usepackage{float}
\usepackage{needspace}
\usepackage[caption=false]{subfig}

\newcommand{\gtlogit}{g_t^{\mathrm{logit}}}

\newcommand{\obsbox}[1]{%
\par\smallskip
\noindent\colorbox{black!5}{%
  \parbox{\dimexpr\linewidth-2\fboxsep\relax}{\textbf{Takeaway.} #1}}%
\par\smallskip
}
\newcommand{\kvsubfloat}[2]{%
\subfloat[#1]{\includegraphics[width=0.48\linewidth]{#2}}%
}

\title{MarginGate: Sparse Margin-Triggered Verification for\\Batch-Invariant LLM Inference}

\author{
Kexin Chu \\
University of Connecticut \\
\texttt{kexin.chu@uconn.edu}
\And
Yang Zhou \\
UC Davis \\
\texttt{yayzhou@ucdavis.edu}
\And
Wei Zhang \\
University of Connecticut \\
\texttt{wei.13.zhang@uconn.edu}
}

\begin{document}
\maketitle

\begin{abstract}
Temperature-zero BF16 LLM inference is often treated as reproducible, yet the same request can emit different tokens when decoded alone or inside a larger batch. Existing fixes use batch-invariant operators or LLM-42's per-token verification, incurring cost even when most steps are stable. We ask whether verification can be applied exclusively to flipped tokens. Across five models, batch-induced token flips are sparse on the flip-rate benchmarks: on MATH500, Llama-3.1-8B flips on $0.48\%$ of synchronous decode steps, and all tested models stay within the $0.3$--$1.3\%$ range on MATH500, GSM8K, and HumanEval. K/V perturbations remain flat before flips, while low top-1/top-2 logit margins expose much of the flip risk. \textbf{MarginGate} turns these observations into a verifier policy: it keeps BF16 decoding on high-margin steps, verifies only low-margin steps, and repairs confirmed mismatches by replacing the current K/V column. We evaluate on four datasets, calibrating on MATH500 and transferring to GSM8K, SharedGPT, and HumanEval. MarginGate restores $100\%$ sequence-level deterministic decoding on Llama-3.1-8B and Qwen2.5-14B with $18.56\%$/$15.05\%$ verifier trigger rates, reducing LLM-42's latency increment by $\mathbf{2.23\times}$/$1.99\times$ relative to always-on verification. On DSR1-Distill-Qwen-7B, the same policy reaches determinism in a harder regime at $49.50\%$ triggers.
\end{abstract}

\section{Introduction}
\label{sec:intro}

Greedy decoding is often used as a reproducible primitive. Evaluation pipelines, LLM-as-a-judge setups, regression tests, distillation loops, and production services commonly assume that a fixed prompt and a fixed decoding policy define a fixed computation~\citep{gu2024minillm,song2025good,zhang2025tpinvariant,llm42}. This assumption fails in BF16 inference: the same request can emit different tokens when decoded alone and when decoded inside a larger batch, even at temperature zero~\citep{song2025good,yuan2025numerical,zhang2025tpinvariant}. The difference is not sampling randomness. It is a numerical effect exposed by the serving shape.

Serving systems make that shape a moving target. High-throughput engines co-schedule requests through dynamic batching, prefill/decode scheduling, paged K/V caches, disaggregated execution, and multi-tier K/V-cache management~\citep{yu2022orca,kwon2023vllm,agrawal2024sarathi,zhong2024distserve,chu2025mcam}. A user usually does not control which other requests share the batch, or when the batch shape changes. From that user's perspective, deterministic behaviour should not depend on whether a request happens to be decoded alone or alongside other prompts.

Existing remedies recover determinism by paying a cost on a broad surface. Batch-invariant operator libraries and LayerCast patch reductions or upcast linear layers across the decode loop~\citep{batchinvariantops2024,yuan2025numerical}. Serving stacks such as vLLM and SGLang have also begun to expose deterministic or batch-invariant paths~\citep{kwon2023vllm,vllmBatchInvariance2026,sglang2024,sglangDeterministicInference2026}. LLM-42~\citep{llm42} takes a different route: it keeps the default BF16 path but verifies every decoded token against a deterministic re-computation and repairs disagreements. These methods establish that deterministic decoding is recoverable. They also raise a narrower question: which decoded steps actually need verification?

Our answer starts from measurement: batch-induced token flips are \emph{rare, local, and signaled by near-tie margins}. Across five open-weight LLMs and the three flip-rate benchmarks, only $0.3$--$1.3\%$ of synchronous decode steps flip. Before the first flip, the protected request's K/V trajectory stays close to a deterministic batch-invariant reference, with no measurable time-axis drift in our traces. The branch point appears at the current decode position's K/V cache entry, which we call the current K/V column. The current step's top-1/top-2 logit margin then gives a reference-free signal for near-tie flip risk.

We propose \textbf{MarginGate}, a sparse invocation policy for an LLM-42-style verifier. Each step first runs the default BF16 batched path. If the current logit margin falls below a calibrated threshold, MarginGate invokes a deterministic verifier for the protected request. If the verifier changes the token, MarginGate overwrites only the current K/V column and emits the verifier token. The trigger is step-local, while the repair action uses the verifier output as the correction target for that step.

On Llama-3.1-8B, MarginGate restores sequence-level deterministic decoding while launching the verifier on $18.56\%$ of synchronous decode steps, compared with LLM-42's $100\%$ per-token verification. Qwen2.5-14B reaches the same determinism at a $15.05\%$ trigger rate. DSR1-Distill-Qwen-7B also reaches deterministic decoding, but its wider tie zone moves the operating point to a denser $49.50\%$ trigger rate. These results frame MarginGate as a calibrated selective-verification policy: its value is largest when low-margin steps are sparse, and the threshold sweep makes each model's cost-determinism frontier explicit.

\paragraph{Contributions.}
\begin{enumerate}
\item \textbf{Observation: sparse, local, near-tied flips} (\S\ref{sec:bg:stable}--\ref{sec:bg:hypothesis}). Across five models and the three flip-rate benchmarks, only $0.3$--$1.3\%$ of synchronous decode steps flip. K/V traces show no measurable pre-flip drift, low top-1/top-2 margins mark many risky steps, and oracle single-column repair confirms that the repair target is local.
\item \textbf{Selective verification as the design target} (\S\ref{sec:lite}). MarginGate keeps high-margin steps on the BF16 fast path, invokes a deterministic verifier on low-margin steps, and repairs only the current K/V column after a confirmed mismatch. Its goal is sequence-level deterministic decoding with far fewer verifier launches than LLM-42, using a threshold selected by a per-model sweep.
\item \textbf{Implementation and measured gains} (\S\ref{sec:eval}). We implement MarginGate and compare it with BF16, Batch-Invariant Ops, LayerCast, and LLM-42. It restores sequence-level determinism while reducing LLM-42's latency increment by $2.23\times$ on Llama-3.1-8B and $1.99\times$ on Qwen2.5-14B, moving deterministic inference closer to the BF16 fast path. DSR1-Distill-Qwen-7B marks a boundary case where the margin trigger becomes dense.
\end{enumerate}
\section{Motivating Observations}
\label{sec:bg}

\subsection{Verification Opportunity}
\label{sec:bg:opportunity}

BF16 greedy decoding can be batch-sensitive because the serving shape changes the reduction plan. A single-request path and a batched path can both be valid CUDA schedules, yet accumulate partial sums in a different order and produce last-bit logit differences~\citep{goldberg1991fp,higham2002accuracy,batchinvariantops2024,yuan2025numerical}. If the leading logits are close, such a perturbation can change the argmax token. We compare the protected request inside a larger batch against a deterministic batch-invariant reference trajectory.

Existing deterministic-inference methods pay for this risk broadly. \emph{Global interventions} such as \texttt{batch\_invariant\_ops}~\citep{batchinvariantops2024} and LayerCast~\citep{yuan2025numerical} patch or upcast broad operator classes. \emph{Per-token verification}, as in LLM-42~\citep{llm42}, keeps the BF16 path but checks every decoded token against a deterministic recomputation. These methods establish that deterministic decoding is recoverable. The question for this section is narrower: which steps actually need intervention?

\subsection{Observation 1: Sparse Token Flips}
\label{sec:bg:stable}

We first measure how often the batch shape changes an emitted token before the decoded trajectories have already separated. For each prompt, we compare the protected request inside a $bs{=}N$ batch with the deterministic reference trajectory. A \emph{token-divergence event} occurs when the two runs emit different tokens at the same position. The denominator is restricted to \emph{synchronous samples}: positions before any earlier token divergence for that trial. This restriction matters because, after the first divergence, the protected request is decoding a different continuation, so later mismatches no longer isolate the local batch-induced perturbation.
We use ``token flip'' as shorthand for this synchronous token-divergence event.

\begin{table}[t]
\centering\small
\begin{tabular}{lrrr}
\toprule
Model & MATH500 & GSM8K & HumanEval \\
\midrule
Llama-3.2-1B  & $0.36\%$ & $0.45\%$ & $0.59\%$ \\
Llama-3.1-8B  & $0.48\%$ & $0.32\%$ & $0.51\%$ \\
Qwen2.5-7B    & $0.67\%$ & $0.86\%$ & $1.27\%$ \\
Qwen2.5-14B   & $0.47\%$ & $0.61\%$ & $0.84\%$ \\
DSR1-Dist.-7B & $1.16\%$ & $1.02\%$ & $1.09\%$ \\
\bottomrule
\end{tabular}
\caption{Token-level non-determinism rate across the batch-size sweep.}
\label{tab:flip_rate}
\end{table}

Table~\ref{tab:flip_rate} reports the divergence rate across five open-weight LLMs and the three flip-rate benchmarks. Rates stay in the \textbf{$0.3$--$1.3\%$ range}: more than $98.7\%$ of synchronous decode steps are token-stable across the tested batch shapes. DSR1-Distill-Qwen-7B stays near $1\%$ across benchmarks, and HumanEval is slightly higher than MATH500 on most models.

This observation exposes redundancy in always-on verification, but it does not yet prove that token-stable steps can be skipped safely. The hidden K/V cache could still drift while the emitted token agrees across batch shapes. The design problem is therefore to find a high-recall at-risk set that contains the rare future divergences while skipping high-margin stable steps.

\obsbox{The observed sparsity motivates selective deterministic intervention: a high-recall at-risk set can reduce verifier use while preserving coverage of potential token flips.}

\subsection{Observation 2: No Pre-Flip K/V Drift}
\label{sec:bg:nocum}

A token being stable across batch shapes does not imply that the underlying \emph{K/V cache state} is stable. If the batched K/V trajectory drifted at token-stable steps, even a good divergence detector would eventually operate on a desynchronized cache. To test this failure mode, we track the protected request's batched K/V trajectory relative to the deterministic reference. For each position $p$, we measure $E^K_p=\|K^{bs=N}_{0,:,p,:}-K^{ref}_{:,p,:}\|_2$, and analogously $E^V_p$, per layer and position over the decoded trajectory. Each trial is aligned to its own first token-divergence position, with $\Delta=p-p_{\mathrm{div}}$.

\begin{figure}[t]
\centering
\includegraphics[width=0.99\linewidth]{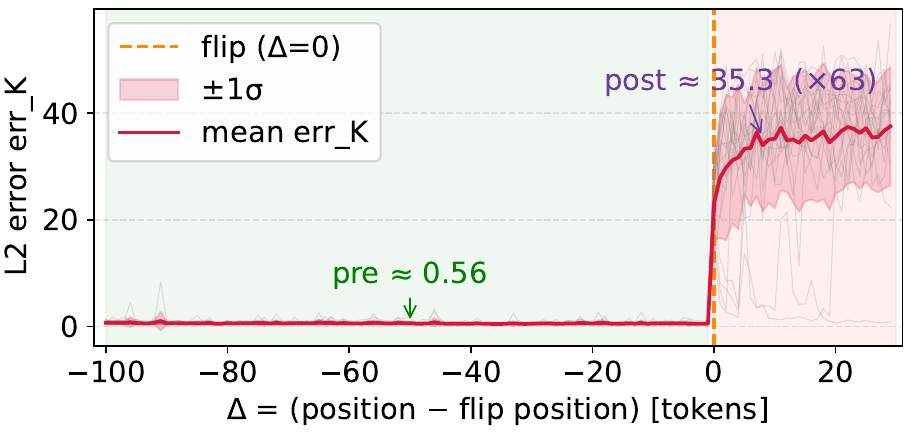}\\[2pt]
\includegraphics[width=0.99\linewidth]{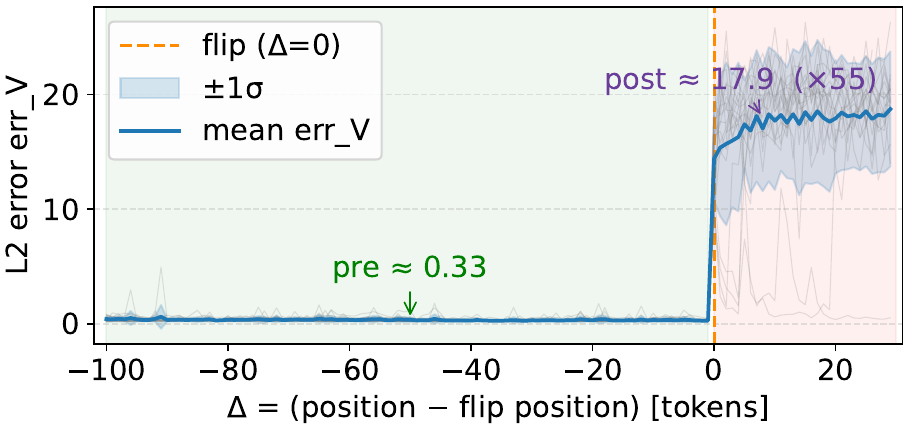}
\caption{Divergence-aligned K/V-cache deviation for the final layer of Llama-3.1-8B, measured relative to the deterministic reference trajectory. Top: K-error; bottom: V-error. Deviations stay near the pre-divergence noise floor for $\Delta<0$ and spike at the first token divergence ($\Delta=0$).}
\label{fig:err_vs_dist}
\end{figure}

Figure~\ref{fig:err_vs_dist} shows the representative Llama-3.1-8B result. Before token divergence, both K and V stay at a flat noise floor. At $\Delta\!=\!0$, the first cross-batch token divergence creates a discontinuous jump ($41\times$ for K, $44\times$ for V). After divergence the error remains high, but the protected request is then decoding a different continuation, so the comparison is no longer evidence of pre-divergence accumulation. The relevant pattern for MarginGate is \textbf{the flat pre-divergence regime}.

Table~\ref{tab:flip_aligned} shows the same pattern across all five models. Final-layer K/V deviation stays at a bounded pre-divergence baseline and spikes at the first cross-batch token divergence. Even the weakest jump is large ($12\times$ for K and $17\times$ for V on Qwen-7B), indicating a discrete branch point rather than gradual cache drift. Appendix~\ref{app:kv_per_model} gives the full per-model trajectories.

\begin{table}[t]
\centering\small
\setlength{\tabcolsep}{3.5pt}
\begin{tabular}{lrrrr}
\toprule
Model (last layer) & K pre & K spike & V pre & V spike \\
\midrule
Llama-1B (L15)     & $0.37$ & $60\times$ & $0.21$ & $82\times$ \\
Llama-8B (L31)     & $0.56$ & $41\times$ & $0.33$ & $44\times$ \\
Qwen-7B (L27)      & $1.37$ & $12\times$ & $4.81$ & $17\times$ \\
Qwen-14B (L47)     & $1.00$ & $28\times$ & $1.93$ & $30\times$ \\
DSR1-Dist.-7B (L27) & $0.72$ & $23\times$ & $2.24$ & $39\times$ \\
\bottomrule
\end{tabular}
\caption{Divergence-aligned final-layer K/V-cache deviation summary at $bs{=}8$, measured relative to the deterministic reference trajectory. ``pre'' is the median pre-divergence mean $\ell_2$ deviation over $\Delta<0$; ``spike'' is the $\Delta=0$ deviation divided by pre.}
\label{tab:flip_aligned}
\end{table}

A per-trial linear regression over the pre-divergence window gives a median deepest-layer K-slope within $|0.007|$/token on all five models, indicating no measurable time-axis drift. The V traces show the same flat pre-divergence regime, albeit with a different baseline scale in Qwen-family models.

For sparse verification, the measurements indicate that skipped token-stable steps do not silently poison the cache history. The repair target for a detected divergence is therefore local to the current K/V column, rather than a growing window of past columns. Appendix~\ref{app:kvstruct} gives the layer-wise view.

\obsbox{K/V deviations remain bounded before token divergence, supporting current-column repair rather than history-wide cache rewrites.}

\subsection{Observation 3: Near-Tie Logits}
\label{sec:bg:logitsig}

Obs~1--2 make sparse verification plausible, but they do not yet give a runtime trigger. At deployment we cannot compare the current decode step against another batch trajectory online; the trigger must come from the protected request's current logits. We therefore inspect the logit vector $\ell$ over the vocabulary at token-divergence events and ask whether risky steps have a reference-free signature.

\textbf{(A) Identity: what token changes across batch shapes?} Table~\ref{tab:logit_topk} first checks whether divergences are near-top swaps or arbitrary vocabulary jumps. We rank the reference token inside the corresponding $bs{=}N$ logits. On the Llama and Qwen instruct models, top-$3$ captures every divergence. DSR1-Distill-Qwen-7B has a wider spread: $88.1\%$ of its divergences are within top-$8$, and about $10\%$ place the alternative token outside top-$50$, consistent with its denser operating point in \S\ref{sec:eval:threshold}.

\begin{table}[t]
\centering\small
\begin{tabular}{l rrr}
\toprule
Model & alt$_{\!\mathrm{tok}}\!\in\!$top-2 & top-3 & top-8 \\
\midrule
Llama-3.2-1B  & $94.8\%$ & $100\%$ & $100\%$ \\
Llama-3.1-8B  & $94.1\%$ & $100\%$ & $100\%$ \\
Qwen2.5-7B    & $98.9\%$ & $100\%$ & $100\%$ \\
Qwen2.5-14B   & $95.8\%$ & $100\%$ & $100\%$ \\
DSR1-Dist.-7B & $69.7\%$ & $77.1\%$ & $88.1\%$ \\
\bottomrule
\end{tabular}
\caption{Cumulative percentage of token-divergence events where the cross-batch alternative token appears in the protected request's top-$K$ logits (synchronous samples). Divergences are concentrated among leading candidates, supporting the view that batch-induced token changes are near-top swaps rather than arbitrary vocabulary jumps.}
\label{tab:logit_topk}
\end{table}

\textbf{(B) Geometry: is the top-1 isolated?} The identity check above still requires two trajectories. For a deployable trigger, we instead ask whether the current logits themselves expose a near tie. Define $N(\Delta)\!=\!|\{j:\ell_j\!\ge\!\ell_{(1)}-\Delta\}|$, the count of logits within $\Delta$ of the top-1. The event $N(\Delta)\!>\!1$ is exactly the condition that the top-1/top-2 margin $\gtlogit\!=\!\ell_{(1)}-\ell_{(2)}$ is below $\Delta$. We compare token-stable and divergent steps at four scales of $\Delta$ in Table~\ref{tab:logit_cluster}.

\begin{table*}[t]
\centering\small
\setlength{\tabcolsep}{2.5pt}
\begin{tabular}{l rrr rrr rrr rrr}
\toprule
 & \multicolumn{3}{c}{$\Delta\!=\!0.25$} & \multicolumn{3}{c}{$\Delta\!=\!0.5$} & \multicolumn{3}{c}{$\Delta\!=\!1.0$} & \multicolumn{3}{c}{$\Delta\!=\!2.0$} \\
Model & stable & div. & ratio & stable & div. & ratio & stable & div. & ratio & stable & div. & ratio \\
\midrule
Llama-3.2-1B  & $1.06$ & $2.21$ & $2.08\times$ & $1.12$ & $2.59$ & $2.31\times$ & $1.25$ & $2.95$ & $2.36\times$ & $1.61$ & $4.81$ & $2.99\times$ \\
Llama-3.1-8B  & $1.07$ & $2.13$ & $2.00\times$ & $1.12$ & $2.24$ & $1.99\times$ & $1.25$ & $2.66$ & $2.13\times$ & $1.54$ & $3.72$ & $2.42\times$ \\
Qwen2.5-7B    & $1.04$ & $1.86$ & $1.80\times$ & $1.07$ & $2.05$ & $1.92\times$ & $1.14$ & $2.34$ & $2.06\times$ & $1.31$ & $3.02$ & $2.31\times$ \\
Qwen2.5-14B   & $1.04$ & $2.17$ & $2.09\times$ & $1.07$ & $2.31$ & $2.16\times$ & $1.14$ & $2.71$ & $2.38\times$ & $1.30$ & $3.65$ & $2.81\times$ \\
DSR1-Dist.-7B & $1.05$ & $1.78$ & $1.70\times$ & $1.09$ & $1.99$ & $1.82\times$ & $1.19$ & $2.29$ & $1.93\times$ & $1.43$ & $3.12$ & $2.18\times$ \\
\bottomrule
\end{tabular}
\caption{Mean $N(\Delta)\!=\!|\{j:\ell_j\!\ge\!\ell_{(1)}-\Delta\}|$ at token-stable vs.\ divergent synchronous steps, and ratio. At stable steps $N\!\approx\!1$ (only the top token is within $\Delta$ of itself); at divergent steps $N\!\approx\!2$ for $\Delta\!\le\!1$, i.e.\ a second near-tied candidate appears. The pattern is consistent across all five models (between $1.7\times$ and $3.0\times$ across all $\Delta$).}
\label{tab:logit_cluster}
\end{table*}

The two tables play complementary roles. Table~\ref{tab:logit_topk} shows that divergences are usually swaps among top candidates, but it is not deployable because it identifies the cross-batch alternative by comparing two trajectories. Table~\ref{tab:logit_cluster} gives the deployable signal: at token-stable steps $N(\Delta)\!\approx\!1$, whereas at divergent steps a second near-tied candidate appears ($N(\Delta)\!\approx\!2$ for $\Delta\!\le\!1$, ratio $1.7$--$3.0\times$ across models). Thus the current logits encode divergence risk through a small top-1/top-2 margin. \S\ref{sec:lite:trigger} uses this signature as the deployable trigger.

\obsbox{The current logit margin provides a deployable risk signal because divergent steps are enriched for near-tied top candidates.}

\subsection{From Observations to MarginGate}
\label{sec:bg:hypothesis}

Obs~1--3 suggest a narrower verifier policy than LLM-42: skip high-margin stable steps, and repair only the current K/V column when a verified token differs. Before relying on the margin trigger end-to-end, we test whether this selective current-column repair would be sufficient if the risky positions were known. The oracle fires at the true divergence positions and writes the deterministic reference K/V column at those positions.

\begin{table}[t]
\centering\small
\setlength{\tabcolsep}{5pt}
\begin{tabular}{lrr}
\toprule
Model & repairs/seq. & seq. det. \\
\midrule
Llama-8B       & $1.00$ & $\mathbf{100\%}$ \\
Qwen-14B       & $0.93$ & $\mathbf{100\%}$ \\
Qwen-7B        & $1.82$ & $\mathbf{100\%}$ \\
DSR1-Dist.-Qwen-7B   & $2.35$ & $\mathbf{100\%}$ \\
\bottomrule
\end{tabular}
\caption{Oracle selective repair across $bs\!\in\!\{2,4,8,16\}$. The oracle fires only at true token-divergence positions and replaces the current K/V column with the deterministic reference column. All four models recover sequence-level determinism while requiring only $0.93$--$2.35$ repairs per sequence.}
\label{tab:oracle_xmodel}
\end{table}

Table~\ref{tab:oracle_xmodel} establishes existence: if the divergence positions are known, fewer than $2.4$ repairs per decoded sequence suffice to recover determinism on every measured model and batch shape. Appendix~\ref{app:oracle_more_trace} shows the corresponding K/V traces. MarginGate replaces the oracle's known positions with the margin trigger from Obs~3.

\section{Design: MarginGate}
\label{sec:lite}

\begin{figure}[t]
\centering
  \includegraphics[width=0.99\linewidth]{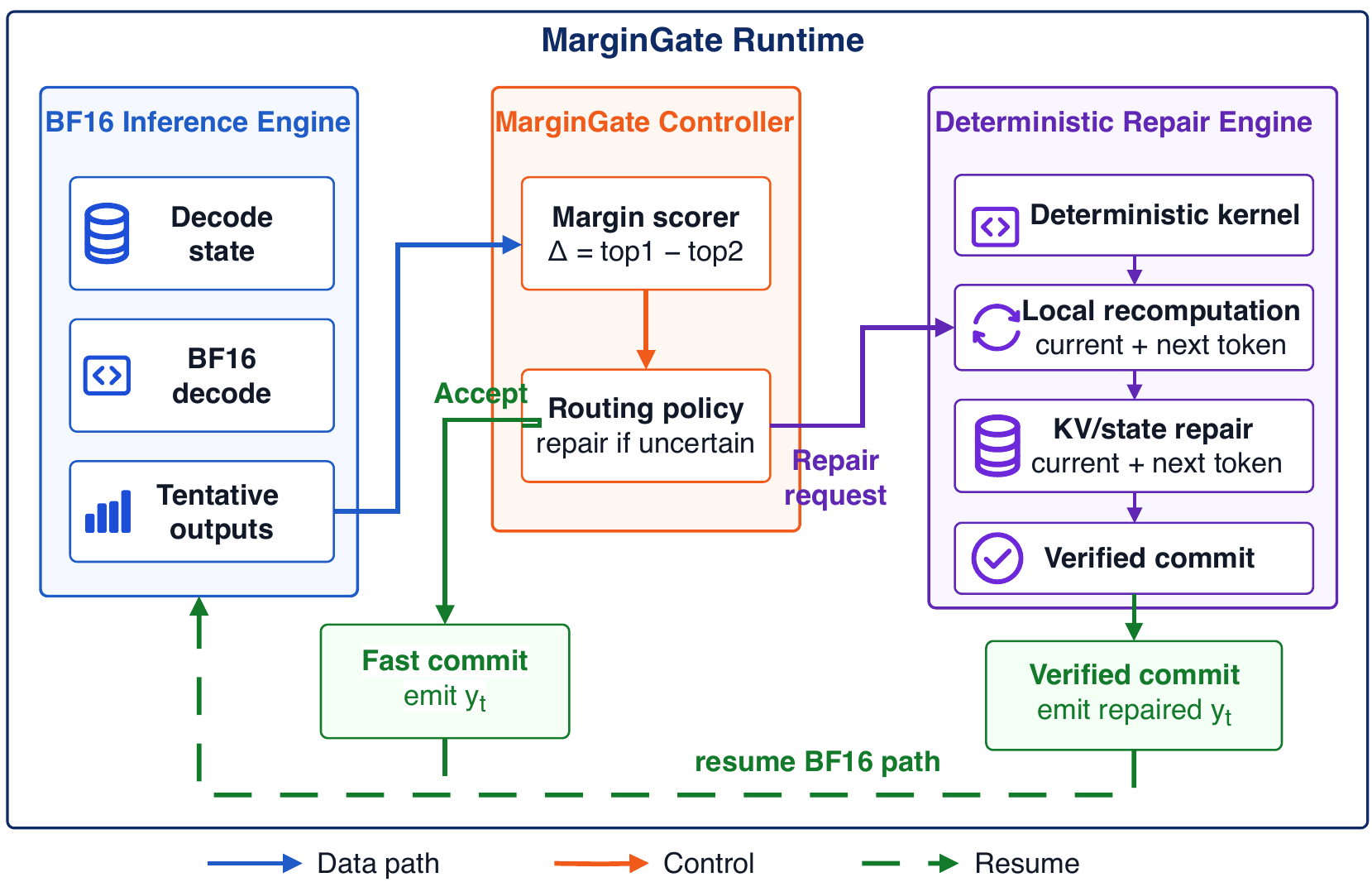}
\caption{MarginGate decode step. The BF16 engine produces tentative outputs. The controller accepts high-margin steps as fast commits. Low-margin steps send a repair request to the deterministic repair engine; verified commits either accept the tentative state or repair the current K/V column.}
\label{fig:margingate_arch}
\end{figure}

MarginGate implements the non-oracle selective verifier from \S\ref{sec:bg:hypothesis}. Its invariant is simple: high-margin steps commit the ordinary batched BF16 state, while low-margin steps ask a deterministic verifier to define the correction target for the current step. If the verifier changes the token, MarginGate overwrites only the current K/V column for the protected request. Figure~\ref{fig:margingate_arch} shows the three runtime components: the BF16 inference engine, the MarginGate controller, and the deterministic repair engine.

\subsection{Trigger and Calibration}
\label{sec:lite:trigger}

\paragraph{Margin trigger.} Obs~3 established the deployable signal. For the protected request $q$, let $\ell_q$ be the logits produced by the default BF16 batched forward at the current step. MarginGate computes only the top two values,
\[
  \gtlogit = \ell_{q,(1)}-\ell_{q,(2)},
\]
and triggers the verifier when $\gtlogit < \tau$. These are the ``uncertain'' steps in Figure~\ref{fig:margingate_arch}. The rule is step-local and reference-free: it reuses logits already produced by the BF16 forward and does not run a second model trajectory unless the trigger fires. The equivalent cluster view from Obs~3, $N(\Delta)\!>\!1$, is the same near-tie test at a chosen scale, so MarginGate implements the top-two margin directly.

\paragraph{Threshold setting.} The threshold $\tau$ is a deployment parameter for each model and serving configuration. The standard argmax perturbation bound~\citep{batchinvariantops2024,yuan2025numerical} supplies a practical starting point: if two batch-shaped logit vectors differ by at most $\varepsilon_{\mathrm{pert}}$ in $\ell_\infty$, their argmaxes can diverge only when the relevant margin is about $2\varepsilon_{\mathrm{pert}}$. We use this value only to choose the sweep range. The final operating point is the smallest tested threshold that reaches sequence-level deterministic decoding on the calibration suite.

\subsection{Single-Column Repair}
\label{sec:lite:action}

The repair action follows from Obs~2 and the oracle experiment in \S\ref{sec:bg:hypothesis}. The BF16 forward first tentatively appends the current token and K/V column. If the margin trigger does not fire, MarginGate performs a \emph{fast commit}: it emits the BF16 token and keeps the tentative cache state. If the trigger fires, the controller sends a repair request to the deterministic repair engine. The verifier recomputes the current transition for the protected request and returns the token plus the K/V column needed to continue the next step. When the verifier and BF16 tokens agree, MarginGate performs a \emph{verified commit} without changing the tentative cache state. When they differ, MarginGate emits the verifier token and copies the verifier-produced K and V for this single column into the protected request's cache. Earlier columns are not rewritten because Obs~2 shows no pre-divergence accumulation.

\paragraph{Deterministic verifier contract.} The verifier must be deterministic for the same model weights, prompt prefix, decoding policy, and fixed deterministic kernels/settings. It does not need to reproduce any particular BF16 serving shape bit-for-bit. At a triggered step, the verifier supplies a repeatable token and K/V column for the current transition; these become the correction target if the BF16 token disagrees. Any verifier satisfying this deterministic contract can be used, including an LLM-42-style recomputation; \S\ref{sec:eval:ablations} tests why the current-column repair action matters.

\subsection{Verifier Accounting}
\label{sec:lite:algo}

MarginGate's cost is determined by two rates. The verifier rate $r_{\mathrm{verify}}=\Pr[\gtlogit<\tau]$ is the fraction of steps that send a repair request in Figure~\ref{fig:margingate_arch}. The repair rate $r_{\mathrm{repair}}$ is the subset where the verifier token disagrees and the current K/V column is overwritten. BF16 has $r_{\mathrm{verify}}=0$, while LLM-42-style verification has $r_{\mathrm{verify}}=1$. Evaluation reports the trigger rate as $r_{\mathrm{verify}}$ because it drives latency; $r_{\mathrm{repair}}$ explains how often MarginGate actually changes the decode state. The selected operating point can be compared with always-on verification under the same cost metric, so selective and always-on operation share a common accounting framework.

Operationally, MarginGate is a per-request policy rather than a replacement for the serving engine. A service can attach it only to requests that require reproducible outputs, while other requests continue on the default BF16 path. The online decision uses the current step's logits and the cached threshold, and launches a verifier only when the gate fires. The verifier then defines the canonical transition for the protected request, while single-column repair keeps the emitted token and K/V state aligned. Thus trigger rate, repair action, and sequence-level determinism describe the cost, intervention, and guarantee of one policy.

\section{Evaluation}
\label{sec:eval}

\subsection{Setup}
\label{sec:eval:setup}

\paragraph{Evaluation protocol.} The core Pareto experiments use MATH500 as the calibration frontier for selecting each model's $\tau_{100}$; transfer checks keep the selected threshold fixed on GSM8K, SharedGPT, and HumanEval, so the end-to-end evaluation spans four datasets~\citep{hendrycks2021math,cobbe2021gsm8k,chiang2023vicuna,chen2021codex}. All runs use greedy decoding and compare the protected request in a batched run against a deterministic batch-invariant reference. Unless otherwise noted, timing and determinism measurements use a $1024$ output-token cap. We report verifier trigger rate, sequence-level determinism, and serving overhead; a run is sequence-deterministic when its complete decoded sequence is identical to the reference. The main method comparison uses Llama-3.1-8B, and the cross-model tables cover five open-weight Llama/Qwen-family models plus DSR1-Distill-Qwen-7B~\citep{llama32,deepseekr1}.

\paragraph{Baselines.} We compare four strategy classes. \emph{BF16 default} is the no-intervention baseline. \emph{Global interventions} such as \texttt{batch\_invariant\_ops}~\citep{batchinvariantops2024} and LayerCast~\citep{yuan2025numerical} patch reductions or linear layers on every step, independent of flip risk. Our Batch-Invariant Ops measurements use the SGLang implementation of the Thinking-Machines kernels. \emph{Always-on verification} (LLM-42~\citep{llm42}) keeps the default path but verifies every token. \emph{Sparse verification} (MarginGate) keeps the LLM-42 repair action but launches it only when $\gtlogit\!<\!\tau$.

\subsection{Main Results}
\label{sec:eval:headline}

The central question is whether MarginGate can recover sequence-level deterministic decoding while avoiding LLM-42's every-token verifier. Table~\ref{tab:walltime_vllm} gives the serving-stack comparison on the same A6000: BF16 is fastest but not deterministic; global Batch-Invariant Ops and LayerCast pay a per-step arithmetic cost; LLM-42 verifies every token; MarginGate verifies only margin-triggered steps and repairs on confirmed disagreement.

\begin{table}[t]
\centering\small
\setlength{\tabcolsep}{3pt}
\resizebox{\columnwidth}{!}{%
\begin{tabular}{l r r r r r}
\toprule
Model & BF16 & Batch-Invariant & LayerCast & LLM-42 & MarginGate \\
 & (s) & (overhead) & (overhead) & (overhead) & (overhead) \\
\midrule
Llama-3.2-1B          & $2.22$ & $+119\%$  & $+133.4\%$ & $+69.9\%$  & $+9.3\%$   \\
Llama-3.1-8B          & $12.27$ & $+143\%$  & $+354.3\%$ & $+64.6\%$  & $+29.0\%$  \\
Qwen2.5-7B            & $12.79$ & $+129\%$  & $+346.9\%$ & $+50.7\%$  & $+49.8\%$ \\
Qwen2.5-14B           & $22.67$ & $+129\%$   & $+373.6\%$ & $+47.5\%$  & $+23.9\%$ \\
DSR1-Dist.-Qwen-7B    & $23.06$ & $+133\%$  & $+345.1\%$ & $+72.0\%$  & $+84.5\%$ \\
\bottomrule
\end{tabular}
}
\caption{End-to-end serving-stack wall-time on A6000 at $bs{=}8$ under the default output-token cap, with LLM-42 using verifier setting $K{=}64$, reported as BF16 latency in seconds and overhead over BF16.}
\label{tab:walltime_vllm}
\end{table}

\paragraph{Implementation details.} Table~\ref{tab:walltime_vllm} measures a public-API prototype. Because the serving APIs used here do not expose mid-generation single-column K/V overwrite, confirmed repairs are implemented by \emph{restart-from-rollback}. This preserves token-level semantics and gives a conservative wall-time measurement; the sweep below therefore reports trigger rate and sequence-level determinism alongside measured overhead.

\subsection{Threshold Operating Points}
\label{sec:eval:threshold}

Before the end-to-end sweep, we estimate the logit perturbation scale on calibration prompts and use $2\!\cdot\!\max(\varepsilon_{\mathrm{pert}})$ only to set a sensible sweep range. The final operating point is selected by the sweep below; Appendix~\ref{app:threshold_calibration} gives the calibration table and shows that DSR1-Distill-Qwen-7B needs a larger empirical threshold because its flip events have wider top-logit spread.

\paragraph{End-to-end sweep.}
\label{sec:eval:pareto}

We then sweep $\tau$ over this calibrated range and run MarginGate end-to-end on the MATH500 calibration suite across the tested batch sizes. Table~\ref{tab:pareto} reports the resulting trigger-rate-vs-determinism frontier, where trigger rate is measured over synchronous decode steps and sequence determinism requires the full decoded sequence to match the deterministic reference output. The main pattern is that the Llama models and Qwen2.5-14B reach deterministic operation with substantially fewer verifier calls than LLM-42, while Qwen2.5-7B and DSR1-Distill-Qwen-7B require denser triggering. The bold rows are the smallest thresholds that make the evaluated runs sequence-deterministic, illustrating MarginGate's intended use as a calibrated frontier rather than a single universal threshold.

\begin{table}[t]
\centering\small
\setlength{\tabcolsep}{3pt}
\resizebox{\columnwidth}{!}{%
\begin{tabular}{l r r r}
\toprule
$\tau$ & trig & seq. det. & overhead \\
\midrule
\multicolumn{4}{l}{\emph{Llama-3.2-1B \quad (LLM-42 $=\!+69.9\%$)}} \\
$0.25$           & $1.22\%$ & $85.0\%$ & $+3.7\%$ \\
$0.5$            & $2.62\%$ & $99.2\%$ & $+4.9\%$ \\
$\mathbf{1.0}$   & $\mathbf{4.90\%}$ & $\mathbf{100.0\%}$ & $\mathbf{+9.3\%}$ \\
$2.0$            & $8.17\%$ & $100.0\%$ & $+16.2\%$ \\
\midrule
\multicolumn{4}{l}{\emph{Llama-3.1-8B \quad (LLM-42 $=\!+64.6\%$)}} \\
$0.5$            & $4.14\%$ & $97.5\%$ & $+6.9\%$ \\
$1.0$            & $7.41\%$ & $98.3\%$ & $+11.9\%$ \\
$2.0$            & $11.75\%$ & $98.3\%$ & $+18.5\%$ \\
$\mathbf{4.0}$   & $\mathbf{18.56\%}$ & $\mathbf{100.0\%}$ & $\mathbf{+29.0\%}$ \\
\midrule
\multicolumn{4}{l}{\emph{Qwen2.5-7B \quad (LLM-42 $=\!+50.7\%$)}} \\
$1.0$            & $7.21\%$ & $75.0\%$ & $+12.8\%$ \\
$2.0$            & $12.07\%$ & $85.8\%$ & $+21.0\%$ \\
$4.0$            & $18.64\%$ & $95.8\%$ & $+32.0\%$ \\
$\mathbf{8.0}$   & $\mathbf{29.15\%}$ & $\mathbf{100.0\%}$ & $\mathbf{+49.8\%}$ \\
\midrule
\multicolumn{4}{l}{\emph{Qwen2.5-14B \quad (LLM-42 $=\!+47.5\%$)}} \\
$1.0$            & $4.97\%$ & $99.2\%$ & $+8.1\%$ \\
$2.0$            & $8.95\%$ & $99.2\%$ & $+14.3\%$ \\
$\mathbf{4.0}$   & $\mathbf{15.05\%}$ & $\mathbf{100.0\%}$ & $\mathbf{+23.9\%}$ \\
$8.0$            & $24.60\%$ & $100.0\%$ & $+38.9\%$ \\
\midrule
\multicolumn{4}{l}{\emph{DSR1-Dist.-Qwen-7B \quad (LLM-42 $=\!+72.0\%$)}} \\
$1.0$            & $12.63\%$ & $48.3\%$ & $+21.3\%$ \\
$2.0$            & $19.69\%$ & $75.0\%$ & $+28.2\%$ \\
$4.0$            & $27.44\%$ & $90.8\%$ & $+37.1\%$ \\
$8.0$   & $37.08\%$ & $97.5\%$ & $+51.8\%$ \\
$\mathbf{16.0}$   & $\mathbf{49.50\%}$ & $\mathbf{100.0\%}$ & $\mathbf{+84.5\%}$ \\
\bottomrule
\end{tabular}
}
\caption{Cross-model MarginGate Pareto frontier on MATH500. ``trig'' is verifier trigger rate; ``seq. det.'' is sequence-level determinism; ``overhead'' follows Table~\ref{tab:walltime_vllm}. Bold rows mark each model's smallest deterministic threshold $\tau_{100}$.}
\label{tab:pareto}
\end{table}

\subsection{Ablations and Robustness}
\label{sec:eval:ablations}

We keep the remaining checks focused on the assumptions behind Table~\ref{tab:pareto}: the margin trigger must identify near-tie steps, the repair action must update the current K/V column after a verifier disagreement, and the calibrated thresholds should transfer beyond homogeneous MATH500 batches.

\paragraph{Repair-action ablation.} A verifier token alone is not enough when the BF16 token disagrees. If the method emits the verifier token but leaves the tentative BF16 K/V column in place, the emitted token and cache state can describe different continuations. MarginGate therefore copies the verifier-produced current K/V column whenever the verifier changes the token.

\paragraph{Heterogeneous batches.} The experiments above use same-prompt-replicated batches to isolate the decode-time batch-invariance effect. We replicate the threshold-trigger check on Llama-3.1-8B with mixed MATH500 prompts per batch, left-padded with per-row position\,$\!=\!0$ at content start to match production serving. This check isolates trigger behavior under mixed prompts. The threshold trigger behaves similarly: synchronous flip rate stays close to the homogeneous run ($0.53\%$ vs.\ $0.48\%$), and at $\tau\!=\!0.5$ the margin trigger recovers all observed flips in the heterogeneous setting. Full numbers are in Appendix~\ref{app:hetero}.

\paragraph{Batch-size scaling.} The serving-stack timings in Table~\ref{tab:walltime_vllm} are measured at $bs{=}8$. Table~\ref{tab:batch_scaling} adds the batch-size sweep for Llama-3.1-8B and Qwen2.5-14B. The purpose is to test how the selected deterministic operating points scale in serving cost as the batch size changes.

\begin{table}[t]
\centering\small
\setlength{\tabcolsep}{5pt}
\begin{tabular}{l r r r}
\toprule
Batch Size & BF16(s) & LLM-42 & MarginGate \\
\midrule
\multicolumn{4}{l}{\emph{Llama-3.1-8B}} \\
$8$    & $12.27$ & $+64.6\%$ & $+29.0\%$ \\
$16$    & $13.10$ & $+69.0\%$ & $+45.4\%$ \\
$32$    & $13.98$ & $+73.8\%$ & $+61.6\%$ \\
$64$   & $16.90$ & $+106.5\%$ & $+73.6\%$ \\
\midrule
\multicolumn{4}{l}{\emph{Qwen2.5-14B}} \\
$8$    & $22.67$ & $+47.5\%$ & $+23.9\%$ \\
$16$    & $23.96$ & $+57.9\%$ & $+36.6\%$ \\
$32$    & $25.96$ & $+65.5\%$ & $+50.0\%$ \\
$64$   & $30.48$ & $+122.7\%$ & $+69.7\%$ \\
\bottomrule
\end{tabular}
\caption{Batch-size scaling on MATH500. Entries report BF16 latency and overhead over BF16; the $bs{=}8$ rows match Table~\ref{tab:walltime_vllm}.}
\label{tab:batch_scaling}
\end{table}

\paragraph{Dataset transfer.} The Pareto sweep in Table~\ref{tab:pareto} uses MATH500. Table~\ref{tab:dataset_transfer} gives the GSM8K, SharedGPT, and HumanEval transfer table under the fixed MATH500-calibrated $\tau_{100}$ values. It reports serving cost under fixed thresholds rather than a new threshold search. LLM-42 is the always-on verifier baseline, while MarginGate keeps the same repair action and changes only when the verifier is invoked.

\begin{table}[t]
\centering\small
\setlength{\tabcolsep}{3pt}
\begin{tabular}{l r r r r}
\toprule
Dataset & BF16(s) & LLM-42 & MarginGate & $\tau$ \\
\midrule
\multicolumn{5}{l}{\emph{Llama-3.1-8B}} \\
MATH500   & $12.27$ & $+64.6\%$ & $+29.0\%$ & $4.0$ \\
GSM8K   & $9.71$ & $+73.3\%$ & $+34.4\%$ & $4.0$ \\
SharedGPT   & $13.38$ & $+44.6\%$ & $+39.4\%$ & $4.0$ \\
HumanEval   & $12.66$ & $+38.3\%$ & $+28.5\%$ & $4.0$ \\
\midrule
\multicolumn{5}{l}{\emph{Qwen2.5-14B}} \\
MATH500   & $22.67$ & $+47.5\%$ & $+23.9\%$ & $4.0$ \\
GSM8K   & $12.04$ & $+49.0\%$ & $+34.5\%$ & $4.0$ \\
SharedGPT   & $19.29$ & $+44.0\%$ & $+40.8\%$ & $4.0$ \\
HumanEval   & $20.67$ & $+32.5\%$ & $+30.3\%$ & $4.0$ \\
\bottomrule
\end{tabular}
\caption{Cross-dataset cost transfer under the MATH500-calibrated threshold. Entries report BF16 latency and overhead over BF16.}
\label{tab:dataset_transfer}
\end{table}

\paragraph{Recall and sequence determinism.} High-recall trigger points can still leave a sequence-level tail because one uncaught flip changes the generated suffix. The K/V traces in \S\ref{sec:bg:nocum} indicate that this tail arises from missed branch points rather than gradual cache drift. The deterministic rows in Table~\ref{tab:pareto} are therefore the operating points used for the cost comparisons.

\FloatBarrier
\section{Discussion}
\label{sec:discussion}

A small top-1/top-2 margin is a risk signal rather than an oracle. Lower thresholds reduce verifier calls, while higher thresholds trade more verifier calls for stronger sequence determinism. Model families with broader near-tie regions, such as DSR1-Distill-Qwen-7B, therefore occupy the denser end of the same frontier. This systems point is narrower than global kernel changes or LayerCast~\citep{batchinvariantops2024,yuan2025numerical}, and orthogonal to K/V-cache tiering, sharing, transport, compression, and quantization~\citep{chu2025mcam,chu2025safekv,liu2024cachegen,liu2024kivi}: MarginGate keeps the verifier-and-repair structure but asks which decode steps need it.

The frontier view gives a deployment rule: choose the smallest threshold that meets a target determinism or latency budget on a held-out calibration stream, then apply it online. Llama and Qwen2.5-14B show a favorable regime where this point is far from always-on verification; DSR1-Distill-Qwen-7B shifts the same curve toward denser verification because near ties are more common. The policy remains calibrated to the model while high-margin steps stay on the BF16 fast path.

Seen this way, MarginGate is a control layer over a serving trajectory. The base engine may be the default BF16 path, a deterministic kernel setting, or a batch-invariant runtime; the policy decides when the protected request consults a deterministic transition. Stronger base kernels can shrink perturbations and move the frontier left, while better risk scores change where verifier calls are placed. The repair rule remains local: after invocation, the verifier token and current K/V column define the request's canonical transition.

\section{Conclusion}
\label{sec:conclusion}

This paper shows that batch-induced token non-determinism in greedy BF16 decoding is often sparse, local, and exposed by near-tie logits. Our measurements find rare token flips, no measurable K/V drift before the first flip, and a step-local margin signal that can trigger verification without running a second trajectory on every token. MarginGate turns these observations into a calibrated verifier policy: it keeps high-margin steps on the ordinary BF16 path, invokes a deterministic verifier on low-margin steps, and repairs only the current K/V column after a confirmed disagreement. On Llama-3.1-8B, this restores sequence-level deterministic decoding while verifying $18.56\%$ of synchronous decode steps instead of every token; Qwen2.5-14B reaches the same guarantee at $15.05\%$. These results suggest that deterministic LLM inference can often be treated as selective verification, keeping most decoding steps on the BF16 fast path while preserving deterministic outputs.

\section*{Limitations}
\label{sec:limitations}

This work focuses on greedy BF16 decode for dense open-weight models. Sampling, beam search, speculative decoding, and other multi-candidate regimes can turn a local token change into different control flow, so they may require different verifier policies~\citep{leviathan2023fast}. The current serving prototype implements confirmed repairs through rollback re-decode because public serving APIs do not expose direct mid-generation K/V-column overwrite; a native runtime integration would remove this extra work. Finally, MarginGate uses a simple top-1/top-2 margin threshold. Since true token flips are sparse, much of the trigger mass is conservative coverage for high recall, and stronger risk scores could further reduce verifier calls.

\bibliography{references}

\clearpage
\appendix
\section{Threshold Calibration Details}
\label{app:threshold_calibration}

We estimate $\varepsilon_{\mathrm{pert}}$ on calibration prompts by comparing the $bs{=}N$ logits with the deterministic reference logits over the protected request's top-$50$ logits. The perturbation-based scale $2\!\cdot\!\max(\varepsilon_{\mathrm{pert}})$ sets the sweep range used in Table~\ref{tab:pareto}; it is not the final threshold selector.
The empirical $\tau_{100}$ is selected only from the MATH500 frontier and is then kept fixed in the transfer and scaling checks. The same cost accounting is used when comparing the selected MarginGate point with always-on verification.

\begin{table}[H]
\centering\small
\setlength{\tabcolsep}{3pt}
\resizebox{\columnwidth}{!}{%
\begin{tabular}{l r r r r}
\toprule
Model & med. $\varepsilon$ & max $\varepsilon$ & pert. $\tau$ & emp. $\tau_{100}$ \\
\midrule
Llama-3.2-1B            & $0.25$ & $0.52$ & $1.03$ & $\phantom{0}1.0$ \\
Llama-3.1-8B            & $0.33$ & $0.91$ & $1.81$ & $\phantom{0}4.0$ \\
Qwen2.5-7B              & $1.66$ & $4.21$ & $8.42$ & $\phantom{0}8.0$ \\
Qwen2.5-14B             & $0.50$ & $1.13$ & $2.26$ & $\phantom{0}4.0$ \\
DSR1-Distill-Qwen-7B    & $0.51$ & $3.88$ & $7.75$ & $16.0$ \\
\bottomrule
\end{tabular}
}
\caption{Threshold calibration used to choose the sweep range. $\varepsilon_{\mathrm{pert}}\!=\!|\ell^{bs=N}\!-\!\ell^{ref}|_\infty$ is measured on the calibration suite; ``pert. $\tau$'' is $2\!\cdot\!\mathrm{max}(\varepsilon_{\mathrm{pert}})$. The empirical $\tau_{100}$ values are the smallest bold operating points in Table~\ref{tab:pareto}.}
\label{tab:eps_pert}
\end{table}

\newpage
\section{Additional K/V Diagnostics}
\label{app:kv_diagnostics}

\subsection{Per-Model Flip-Aligned Trajectories}
\label{app:kv_per_model}

\begin{figure}[H]
\centering
\kvsubfloat{Llama-3.2-1B / K-error}{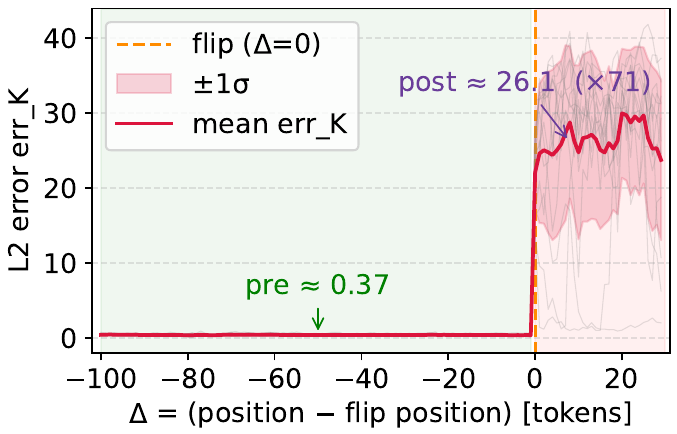}\hfill
\kvsubfloat{Llama-3.2-1B / V-error}{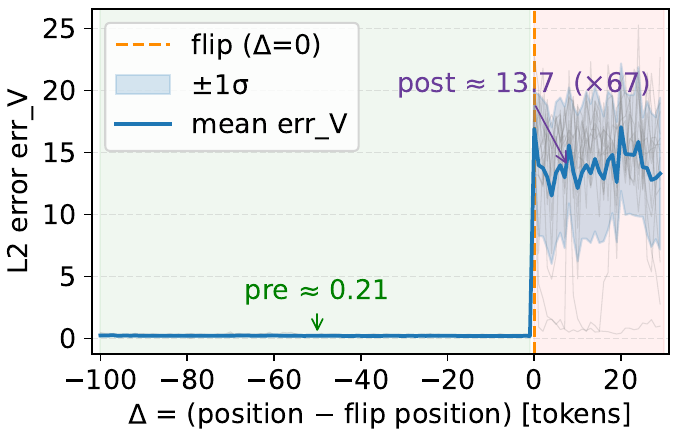}\\[1pt]
\kvsubfloat{Qwen2.5-7B / K-error}{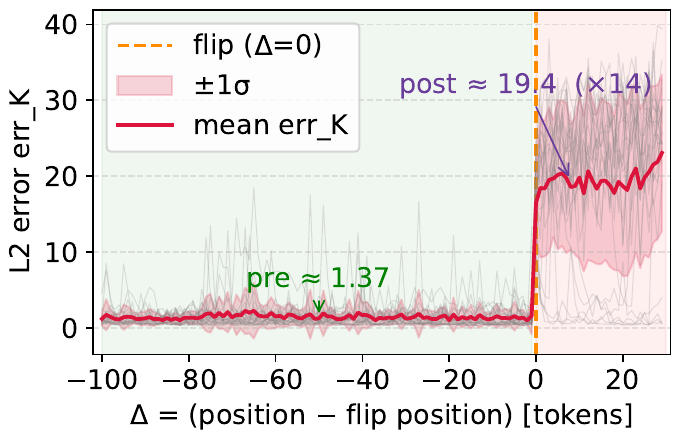}\hfill
\kvsubfloat{Qwen2.5-7B / V-error}{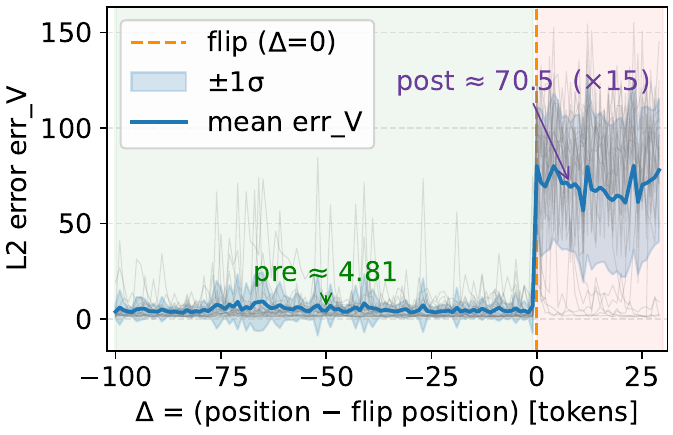}\\[1pt]
\kvsubfloat{Qwen2.5-14B / K-error}{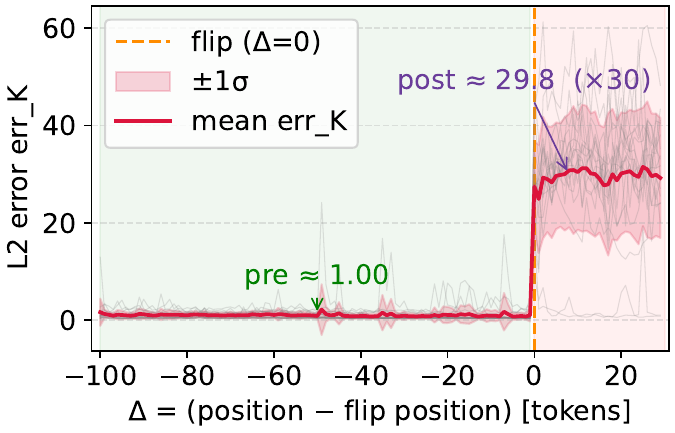}\hfill
\kvsubfloat{Qwen2.5-14B / V-error}{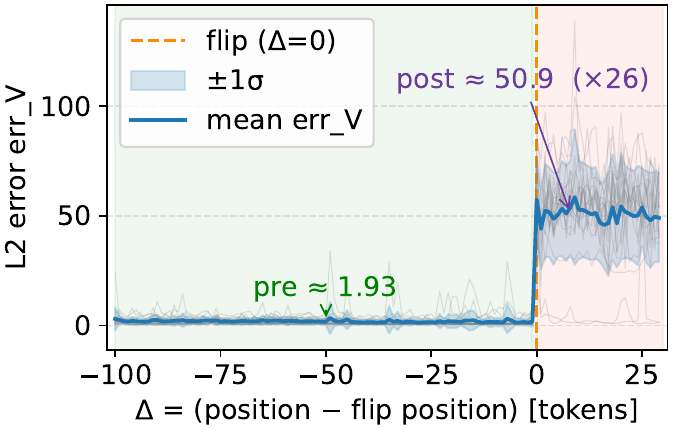}\\[1pt]
\kvsubfloat{DSR1-Distill-Qwen-7B / K-error}{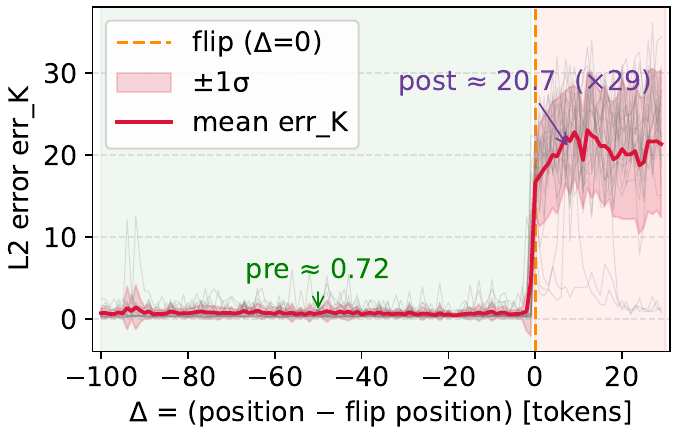}\hfill
\kvsubfloat{DSR1-Distill-Qwen-7B / V-error}{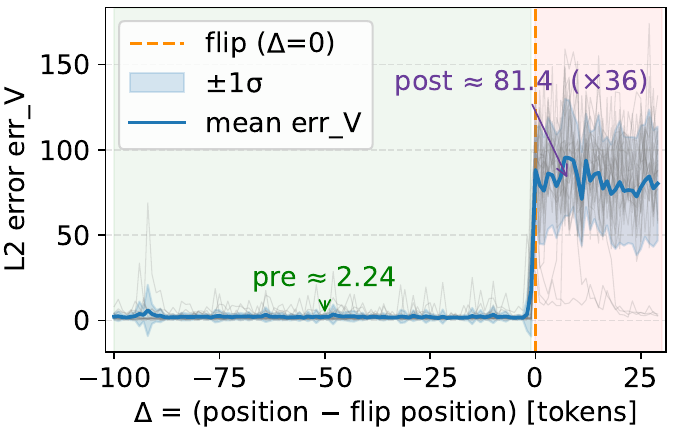}
\caption{Flip-aligned final-layer K/V-cache error trajectories for the four non-representative models, complementing Figure~\ref{fig:err_vs_dist}. Rows correspond to Llama-3.2-1B, Qwen2.5-7B, Qwen2.5-14B, and DSR1-Distill-Qwen-7B; columns show K-error and V-error. Each plot uses the same $bs{=}8$ diagnostic setting. Across model family and scale, both K and V remain at the pre-flip noise floor until $\Delta\!=\!0$, where the first token flip creates a single-column jump.}
\label{fig:app_kv_per_model}
\end{figure}

Figure~\ref{fig:app_kv_per_model} aggregates the flip-aligned K (left) and V (right) error trajectories at the last layer of each of the remaining four models in our cross-model verification. All four model rows match Figure~\ref{fig:err_vs_dist}: flat at $O(\varepsilon)$ pre-flip, a large jump at $\Delta\!=\!0$, and elevated error thereafter (the post-flip continuation is different). The post/pre jump ratios for K/V are $71/65\times$ (Llama-3.2-1B), $14/15\times$ (Qwen2.5-7B), $30/26\times$ (Qwen2.5-14B), and $29/36\times$ (DSR1-Distill-Qwen-7B), with V generally showing the larger baseline in the Qwen-family models.

\FloatBarrier

\subsection{Oracle Repair Traces}
\label{app:oracle_more_trace}

\begin{figure}[H]
\centering
\includegraphics[width=0.99\linewidth]{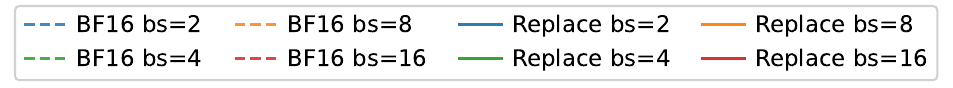}\\[-1pt]
\kvsubfloat{Llama-3.1-8B / K trace}{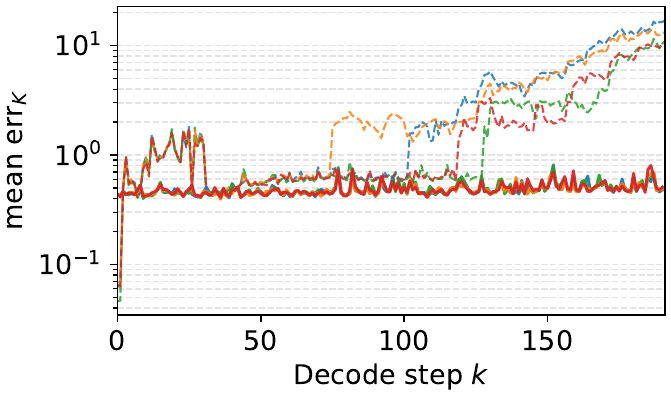}\hfill
\kvsubfloat{Llama-3.1-8B / V trace}{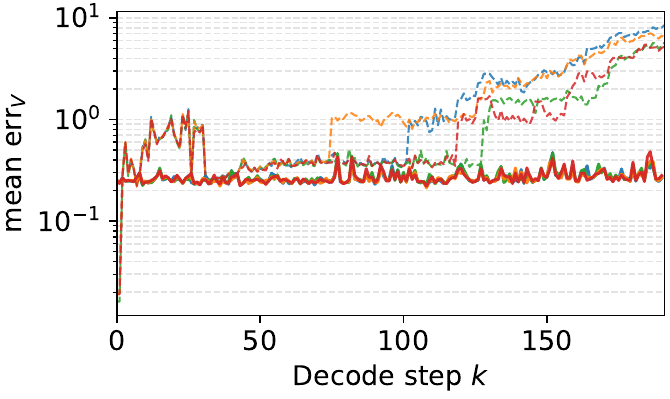}\\[1pt]
\kvsubfloat{Qwen2.5-7B / K trace}{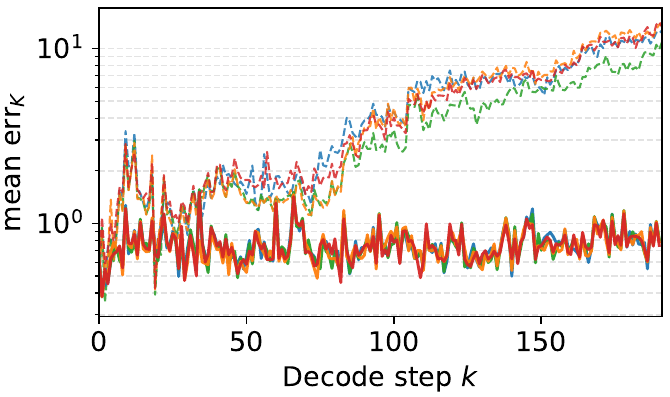}\hfill
\kvsubfloat{Qwen2.5-7B / V trace}{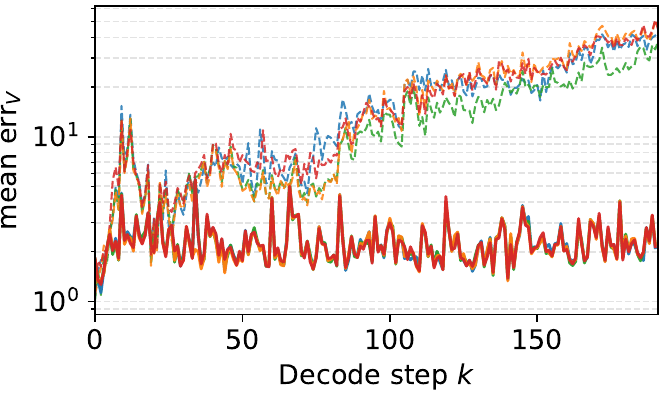}\\[1pt]
\kvsubfloat{Qwen2.5-14B / K trace}{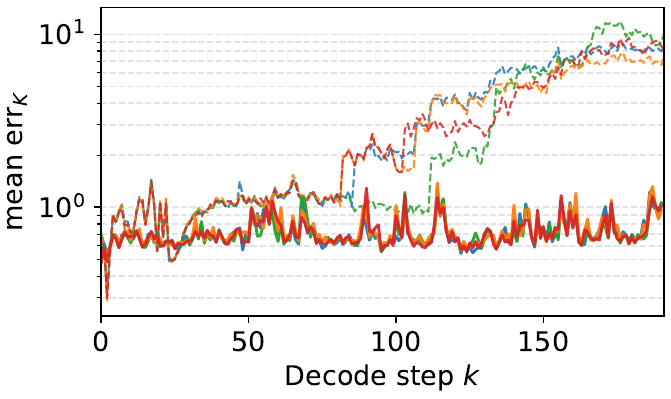}\hfill
\kvsubfloat{Qwen2.5-14B / V trace}{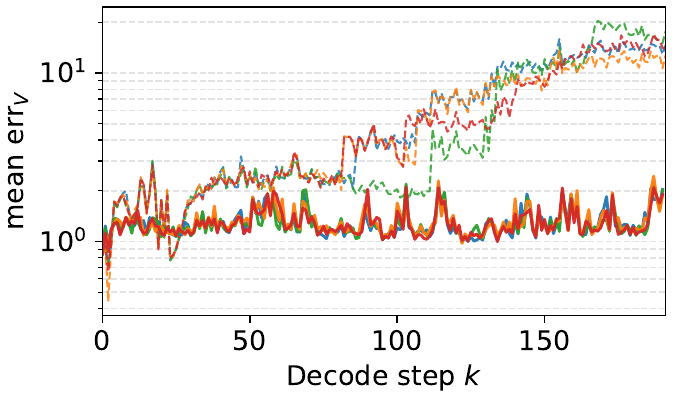}\\[1pt]
\kvsubfloat{DSR1-Distill-Qwen-7B / K trace}{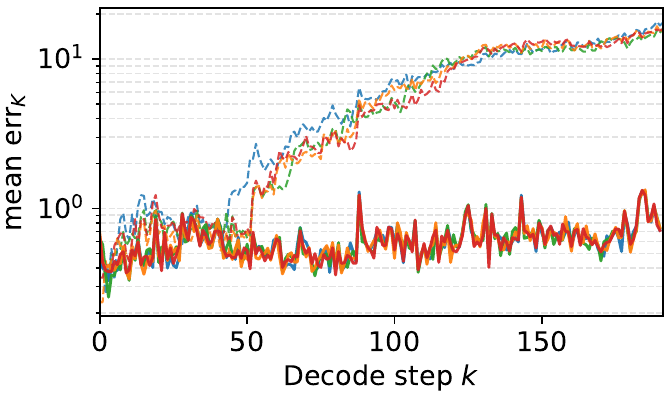}\hfill
\kvsubfloat{DSR1-Distill-Qwen-7B / V trace}{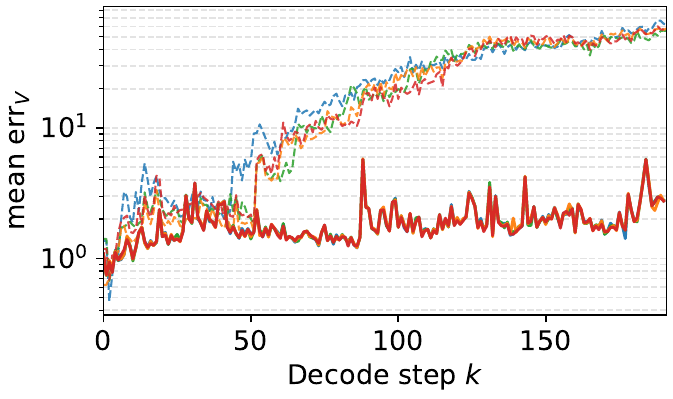}
\caption{Oracle single-column repair traces across the decoded trajectory and four batch sizes (log $y$ axis). Rows correspond to Llama-3.1-8B, Qwen2.5-7B, Qwen2.5-14B, and DSR1-Distill-Qwen-7B; columns show mean err$_K$ and mean err$_V$. Solid curves apply oracle repair; dashed curves are the unrepaired BF16 trajectories.}
\label{fig:oracle_trace_more_models}
\end{figure}

Figure~\ref{fig:oracle_trace_more_models} shows the oracle single-column repair traces for the full replace-intervention models. The oracle uses true divergence positions to isolate the repair action. The unrepaired run drifts after the first flip, while the repaired K and V traces remain at the pre-flip noise floor across all four batch sizes.

\FloatBarrier

\Needspace{16\baselineskip}
\subsection{Layer-Wise Structure}
\label{app:kvstruct}

\begin{figure}[H]
\centering
\includegraphics[width=0.49\linewidth]{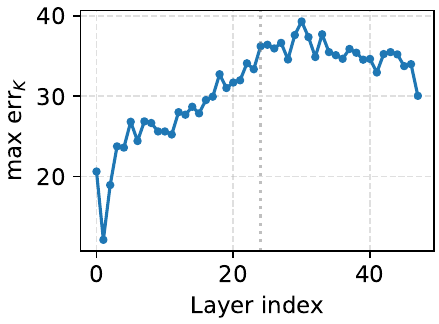}\hfill
\includegraphics[width=0.49\linewidth]{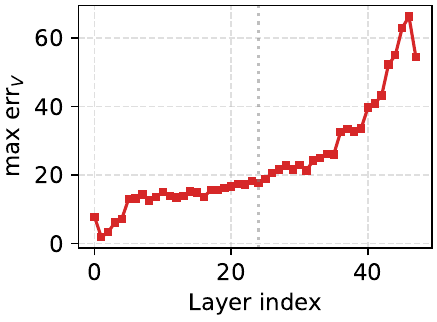}
\caption{Qwen2.5-14B: max-err per layer for K (left) and V (right), aggregated over the same $bs{=}8$ traces. K rises until mid-network then contracts; V is monotonic in depth.}
\label{fig:kv_per_layer}
\end{figure}

K and V cache errors are not symmetric in layer depth. Table~\ref{tab:kv_struct} reports max error at each model's first, middle, and last layer; Figure~\ref{fig:kv_per_layer} shows the same trend at every layer for Qwen2.5-14B.

\begin{table}[H]
\centering\small
\setlength{\tabcolsep}{2pt}
\begin{tabular}{l rrr rrr}
\toprule
 & \multicolumn{3}{c}{\textbf{K-error}} & \multicolumn{3}{c}{\textbf{V-error}} \\
\cmidrule(lr){2-4}\cmidrule(lr){5-7}
Model & first & mid & last & first & mid & last \\
\midrule
Llama-3.2-1B    & $17.0$ & $25.5$ & $20.1$ & $1.4$ & $5.2$ & $10.5$ \\
Llama-3.1-8B    & $29.6$ & $42.6$ & $34.6$ & $1.3$ & $9.0$ & $17.3$ \\
Qwen2.5-7B      & $24.7$ & $39.6$ & $33.8$ & $7.7$ & $20.3$ & $126.3$ \\
Qwen2.5-14B     & $20.6$ & $36.3$ & $30.1$ & $7.8$ & $17.7$ & $54.5$ \\
DSR1-Dist.-7B   & $21.0$ & $35.4$ & $28.8$ & $8.2$ & $22.6$ & $123.6$ \\
\bottomrule
\end{tabular}
\caption{Per-layer max-err of K and V (mean over the diagnostic $bs{=}8$ traces) at each model's first ($L_0$), middle, and last layer. K-error peaks at mid-depth and contracts back toward the output; V-error grows monotonically along depth (last/first ratio $2.6$--$16.4\times$, especially aggressive on the Qwen family).}
\label{tab:kv_struct}
\end{table}

\textbf{Mechanism.} The different K/V depth trends are consistent with their roles in attention: K perturbations affect the softmax scores, while V perturbations flow additively into the residual stream and can grow with layer depth. This helps explain the flip-step spikes summarized in Table~\ref{tab:flip_aligned}. Even though V grows $3$--$13\times$ across depth, per-position V error in the pre-flip trace remains $O(\varepsilon)$ (\S\ref{sec:bg:nocum}), so error does not accumulate with decode step.
\FloatBarrier

\newpage
\section{Heterogeneous Batches}
\label{app:hetero}

The heterogeneous check uses mixed MATH500 batches, with left padding and per-row position IDs so that each content token starts at position $0$. We apply the same synchronous-sample definition as in \S\ref{sec:bg:stable}. This appendix reports the trigger-signal check under mixed prompts; the serving-stack repair measurements use homogeneous batches.

\begin{table}[H]
\centering\small
\setlength{\tabcolsep}{4pt}
\begin{tabular}{l rr}
\toprule
Metric & Homogeneous & Heterogeneous \\
\midrule
Synchronous flip rate     & $0.48\%$   & $0.53\%$ \\
Recall @ $\tau\!=\!0.25$  & $99\%$     & $95\%$ \\
Recall @ $\tau\!=\!0.5$   & $100\%$    & $100\%$ \\
\bottomrule
\end{tabular}
\caption{Llama-3.1-8B threshold trigger on heterogeneous MATH500 batches with per-row position IDs. Behaviour matches the homogeneous-batch evaluation.}
\label{tab:hetero}
\end{table}

\end{document}